\documentclass[10pt,twocolumn,letterpaper]{article}

\usepackage{iccv}
\usepackage{times}
\usepackage{epsfig}
\usepackage{graphicx}
\usepackage{amsmath}
\usepackage{amssymb}

\usepackage[pagebackref=true,breaklinks=true,letterpaper=true,colorlinks,bookmarks=false]{hyperref}
\usepackage[font=small]{caption}
\iccvfinalcopy 

\newcommand{\gray}[1]{\textcolor{gray}{{#1}}}

\newcommand{\blue}[1]{\color[HTML]{3166FF}{{#1}}}
\newcommand{\red}[1]{\color[HTML]{C41E3A}{{#1}}}
\newcommand{\white}[1]{\color[HTML]{FFFFFF}{{#1}}}

\newcommand{\figref}[1]{Fig.~\ref{#1}}

\newcommand{\tabref}[1]{Table~\ref{#1}}
\newcommand{\eqnref}[1]{Eq.~(\ref{#1})}
\newcommand{\secref}[1]{Sec.~\ref{#1}}
\renewcommand{\paragraph}[1]{\vspace{1mm}\noindent\textbf{#1}}
\usepackage{multirow}
\usepackage{tabu}

\usepackage[capitalize]{cleveref}
\crefname{section}{Sec.}{Secs.}
\Crefname{section}{Section}{Sections}
\Crefname{table}{Table}{Tables}
\crefname{table}{Tab.}{Tabs.}

\usepackage{multirow}
\usepackage[table,xcdraw]{xcolor}

\makeatletter
\@namedef{ver@everyshi.sty}{}
\makeatother
\usepackage{tikz}

\usepackage{subcaption}

\usepackage{tikz}
\usetikzlibrary{arrows,intersections}
\tikzset{font=\scriptsize}
\usepackage{pgfplots}
\pgfplotsset{compat=1.11}
\usepgfplotslibrary{fillbetween}
\usetikzlibrary{backgrounds}

\usepackage{xspace}
\usepackage{tabu}
\newlength\savewidth\newcommand\shline{\noalign{\global\savewidth\arrayrulewidth
  \global\arrayrulewidth 1pt}\hline\noalign{\global\arrayrulewidth\savewidth}}
\newcommand{\tablestyle}[2]{\setlength{\tabcolsep}{#1}\renewcommand{\arraystretch}{#2}\centering\footnotesize}

\newcommand{\ours}{CFM-ViT\xspace}

\makeatletter
\newcommand{\thickhline}{%
    \noalign {\ifnum 0=`}\fi \hrule height 1pt
    \futurelet \reserved@a \@xhline
}
\makeatother

\usepackage{bm}
\include{math_commands}

\ificcvfinal\fi

\begin{document}

\title{Contrastive Feature Masking Open-Vocabulary Vision Transformer}

\author{
Dahun Kim  \quad\quad\quad Anelia Angelova \quad\quad\quad Weicheng Kuo\\\\
Google DeepMind\\
}

\maketitle
\ificcvfinal\thispagestyle{empty}\fi

\begin{abstract}
We present Contrastive Feature Masking Vision Transformer (\ours) - an image-text pretraining methodology that achieves simultaneous learning of image- and region-level representation for open-vocabulary object detection (OVD). Our approach combines the masked autoencoder (MAE) objective into the contrastive learning objective to improve the representation for localization tasks.  Unlike standard MAE, we perform reconstruction in the joint image-text embedding space, rather than the pixel space as is customary with the classical MAE method, which causes the model to better learn region-level semantics. Moreover, we introduce Positional Embedding Dropout (PED) to address scale variation between image-text pretraining and detection finetuning by randomly dropping out the positional embeddings during pretraining. PED improves detection performance and enables the use of a frozen ViT backbone as a region classifier, preventing the forgetting of open-vocabulary knowledge during detection finetuning. On LVIS open-vocabulary detection benchmark, \ours achieves a state-of-the-art 33.9 AP$r$, surpassing the best approach by 7.6 points and achieves better zero-shot detection transfer.
Finally, \ours acquires strong image-level representation, outperforming the state of the art on 8 out of 12 metrics on zero-shot image-text retrieval benchmarks. 

\end{abstract}
\vspace{-4mm}

\section{Introduction}

The ability to detect a vast array of objects in the real world is fundamental to computer vision and machine learning. This powers a wide range of applications from autonomous agents to search engines. Unfortunately, to date most modern object detectors rely on manually annotated regions and class labels, which is labor-intensive and impractical to scale beyond the order of $10^3$ categories.

\begin{figure}
\includegraphics[width=\linewidth]{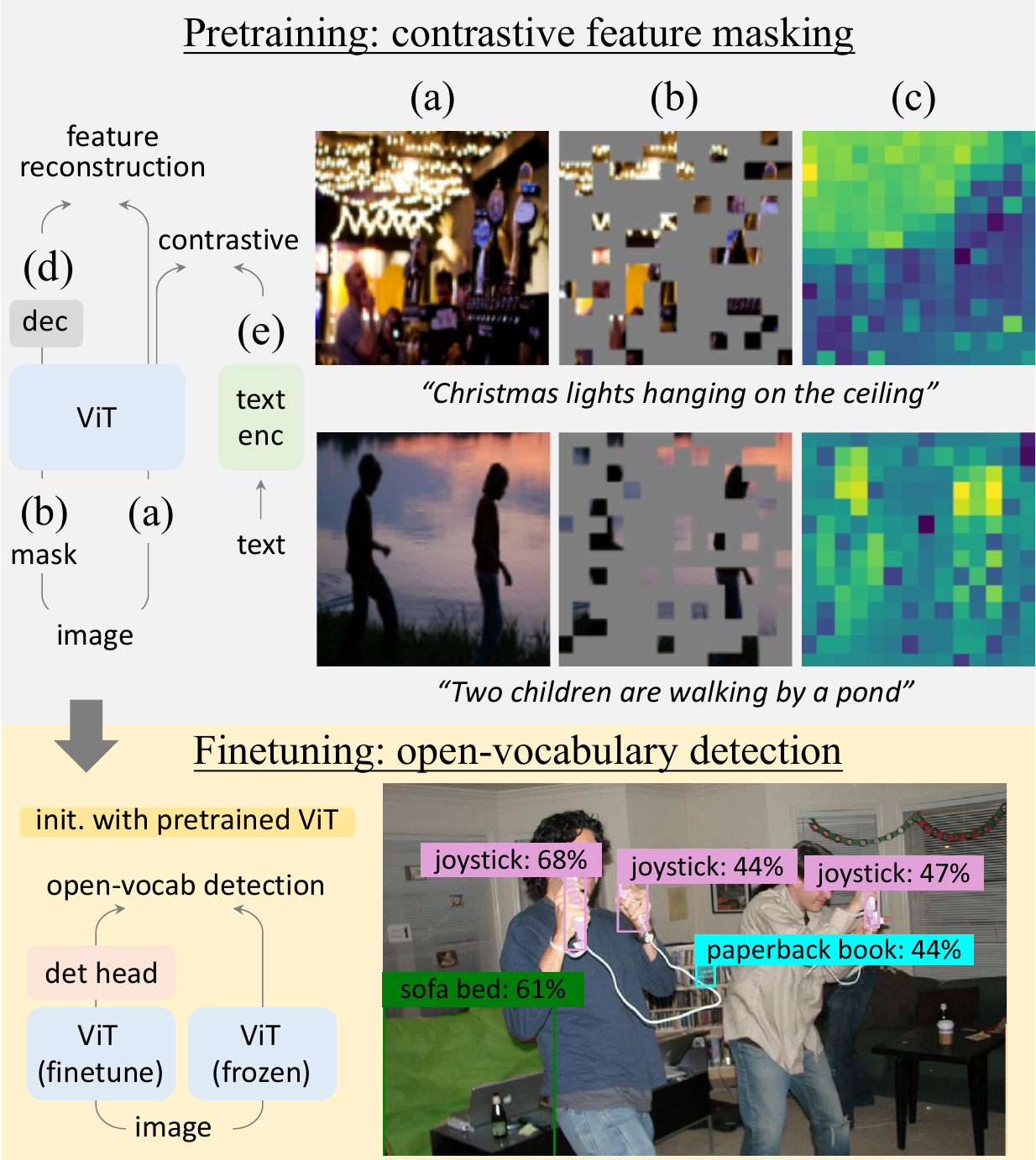}
\caption{We propose \ours to pretrain vision transformers to capture more pixel and region information for open-vocabulary detection. \ours predicts masked contrastive features on top of the contrastive image-text pretraining. (Top) We visualize (c) the similarity map between (d) the \textit{reconstructed} image features (see top left) and (e) the query text embedding. \ours correctly predicts the (c) whole-image semantics from (b) heavily truncated images. (Bottom) Our open-vocabulary detector exploits the frozen ViT backbone to retain pretrained knowledge and is able to detect base and novel object classes (only novel classes are shown).}
\label{fig:teaser}
\vspace{-5mm}
\end{figure}

A new task called open-vocabulary detection (OVD) has been introduced to address the vocabulary limitation in object detection by using image-text pairs for training and text queries from users at test time~\cite{Zareian_2021_CVPR}. Open-vocabulary detectors represent categories as text embeddings rather than discrete class labels, allowing them to predict objects unavailable during training. Various techniques, such as knowledge distillation~\cite{gu2022openvocabulary,du2022learning}, weak supervision~\cite{zhou2022detecting}, self-training~\cite{zhong2021regionclip,rasheed2022bridging,zhao2022exploiting}, and frozen backbone~\cite{kuo2022f}, have been suggested. Typically, CNN backbones are utilized in these approaches. As vision transformers have gained significant traction in image understanding~\cite{dosovitskiy2020image,zhang2022segvit,He_2022_CVPR,bao2021beit}, it is crucial to explore open-vocabulary detectors based on vision transformers~\cite{minderer2022simple}. Moreover, to our knowledge, most current OVD research assumes the availability of pretrained Vision-Language Models (VLMs) (\eg CLIP~\cite{radford2021clip}), and proposes adaptation or finetuning techniques to overcome the disparity between image-level pretraining and object-level finetuning~\cite{gu2022openvocabulary,du2022learning,zhong2021regionclip,zhao2022exploiting,rasheed2022bridging}. However, as these VLMs are typically optimized for image-level tasks such as classification and retrieval, they do not adequately utilize the pixel- and region-level information during pretraining, which is crucial for downstream open-vocabulary detection.

We present \ours (Contrastive Feature Masking Vision Transformer), a simple framework to pretrain vision transformers to capture more detailed pixel/region information for open-vocabulary object detection (\figref{fig:teaser}). Inspired by MAE~\cite{He_2022_CVPR}, we adopt the concept of masked auto-encoding to enhance object representation during pretraining. However unlike MAE, we perform prediction in the joint image-text embedding space rather than the pixel space as an auxiliary objective to the contrastive image-text learning. This additional objective provides orthogonal signal from the contrastive learning, and benefits downstream detection task without compromising the image-level tasks. In addition, we propose Positional Embedding Dropout (PED) to address overfitting to the typically lower-resolution and object-centric pretraining data. By randomly dropping out positional embeddings during pretraining, PED aids the model to learn more robust representations that better generalize to high-res detection data. Moreover, PED enables the use of a frozen ViT encoder as an open-vocabulary region-classifier, which prevents the forgetting of open-vocabulary knowledge at detection.

We evaluate \ours on the widely used LVIS and COCO open-vocabulary detection benchmarks. Our top-performing model obtains 33.9 AP$r$ on LVIS, surpassing the previous best approach by 7.6 AP$r$ at system level. 
On the COCO benchmark, \ours represents the first ViT-based model and achieves a very competitive novel AP without using pseudo labels or weak supervision. 
Although not optimized for retrieval, \ours outperforms the state-of-the-art methods of similar or larger capacity on 8 out of 12 image-text retrieval benchmark metrics. 
In summary:
\begin{itemize}
    \item We present an image-text pretraining methodology (\ours) to learn localization cues for open-vocabulary detection by contrastive feature masking.
    \item We propose Positional Embedding Dropout (PED) to bridge the gap between image-text pretraining and detection finetuning, which enables the use of a frozen ViT encoder to prevent the forgetting of open-vocabulary knowledge during detection finetuning.
    \item \ours achieves state-of-the-art AP$r$ on LVIS open-vocabulary detection benchmark, shows very competitive performance on COCO and zero-shot transfer to Objects365, and outperforms the SOTA on 8 out of 12 metrics of zero-shot image-text retrieval benchmarks.
\end{itemize}
We hope these discoveries would encourage the community to explore open-vocabulary detection from the perspective of image-text pretraining.

\section{Related Works}
\label{sec:related}
\paragraph{Language-supervised open-vocabulary recognition.}\quad
Learning representation for open-vocabulary recognition is a hallmark of general intelligence. Early pioneering works such as DeViSE~\cite{frome2013devise} and ConSE~\cite{norouzi2013zero} used deep convolutional networks to construct a shared image-text embedding space for zero-shot recognition. To leverage the co-occurrence of image and text in raw internet data, researchers have explored various data sources such as image tags~\cite{chen2015webly,divvala2014learning,joulin2016learning}, captions~\cite{desai2021virtex,he2017fine,sariyildiz2020learning,wang2009learning}, alt-texts~\cite{align, schuhmann2021laion}, image search queries~\cite{radford2021clip}, page title~\cite{chen2022pali}, or a combination of these sources~\cite{chen2022pali}. From a modeling perspective, contrastive learning has become a popular paradigm because of its simplicity, scalability, and versatility in zero-shot, few-shot, and full finetuning transfer settings~\cite{basic, radford2021clip, li2022scaling,dong2022clip,pmlr-v162-li22n}. While most of these works focus on image-level understanding, we explore the learning of region-level information in the image-text pretraining, which is essential for open-vocabulary detection task.

\paragraph{Self-supervised object representation learning.}\quad
Scaling up annotation for detection presents a significant challenge. As a result, many efforts have been made to learn object representations in a self-supervised manner. These approaches can be broadly categorized as contrastive or generative. 
These contrastive approaches typically use sliding windows~\cite{Xiao_2021_ICCV}, object proposals~\cite{SoCo,Henaff_2021_ICCV}, or point samples~\cite{Bai_2022_CVPR} for pixel or region-level contrastive learning. Generative methods use masked image modeling with reconstruction targets such as pixels ~\cite{He_2022_CVPR}, low-level~\cite{bao2021beit,wei2022masked} / high-level image features~\cite{chen2022sdae,zhou2021ibot}, or combine with the contrastive objective~\cite{huang2022contrastive}. By learning to restore masked images, the model needs to learn about objects and regions. However, although these self-supervised methods are suited for localization tasks, they lack the necessary image-text learning for open-vocabulary recognition. Some recent works~\cite{wei2022mvp,peng2022beit,zhang2022cae,hou2022milan,fang2022eva} utilize off-the-shelf CLIP features~\cite{radford2021clip} as prediction targets to enhance masked image modeling by two-stage training. In this work, we propose a novel approach to combine generative self-supervised learning jointly with contrastive image-text learning in a single end-to-end training stage. While some concurrent works have explored similar objectives for zero-shot image-level tasks or fully supervised finetuning~\cite{dong2022maskclip,yang2023masked,su2022towards}, our focus is on open-vocabulary detection.

\paragraph{Open-vocabulary object detection and segmentation.}\quad
Zero-shot detection aims to enhance detection models beyond their limited training categories by aligning region visual representation and category word embeddings~\cite{bansal2018zero, rahman2020improved, demirel2018zero, zheng2020background} or generating visual features with a generative model~\cite{hayat2020synthesizing,zhu2020don}. Open-vocabulary detection~\cite{Zareian_2021_CVPR} improves upon zero-shot detection by incorporating image-text supervision about the novel categories. With the advent of image-text pretraining, numerous studies have explored adapting these pretrained models to open-vocabulary detection and segmentation ~\cite{gu2022openvocabulary,zhong2021regionclip,ghiasi2022scaling,li2022language,zhou2022maskclip}. For instance, ViLD~\cite{gu2022openvocabulary} distills image-text knowledge into the detector, while DetPro~\cite{du2022learning} improves ViLD by category prompt optimization. Additionally, region-text self-training has been demonstrated on image caption data~\cite{zhong2021regionclip}, classification data~\cite{rasheed2022bridging}, and unlabeled data~\cite{zhao2022exploiting}. Phrase grounding~\cite{li2021grounded}, weak supervision~\cite{zhou2022detecting}, and frozen model~\cite{kuo2022f} approaches have also been explored. Most methods rely on CNN backbones, but vision transformers are gaining momentum~\cite{minderer2022simple,zhou2022maskclip,kim2023region,kuo2023mammut,li2023reclip}. While previous studies have focused on finetuning or adaptation strategies for pretrained models, ours seeks to improve the image-text pretraining by predicting the masked representation of vision transformer.

\begin{figure*}[t]
\centering
\includegraphics[width=0.97\linewidth]{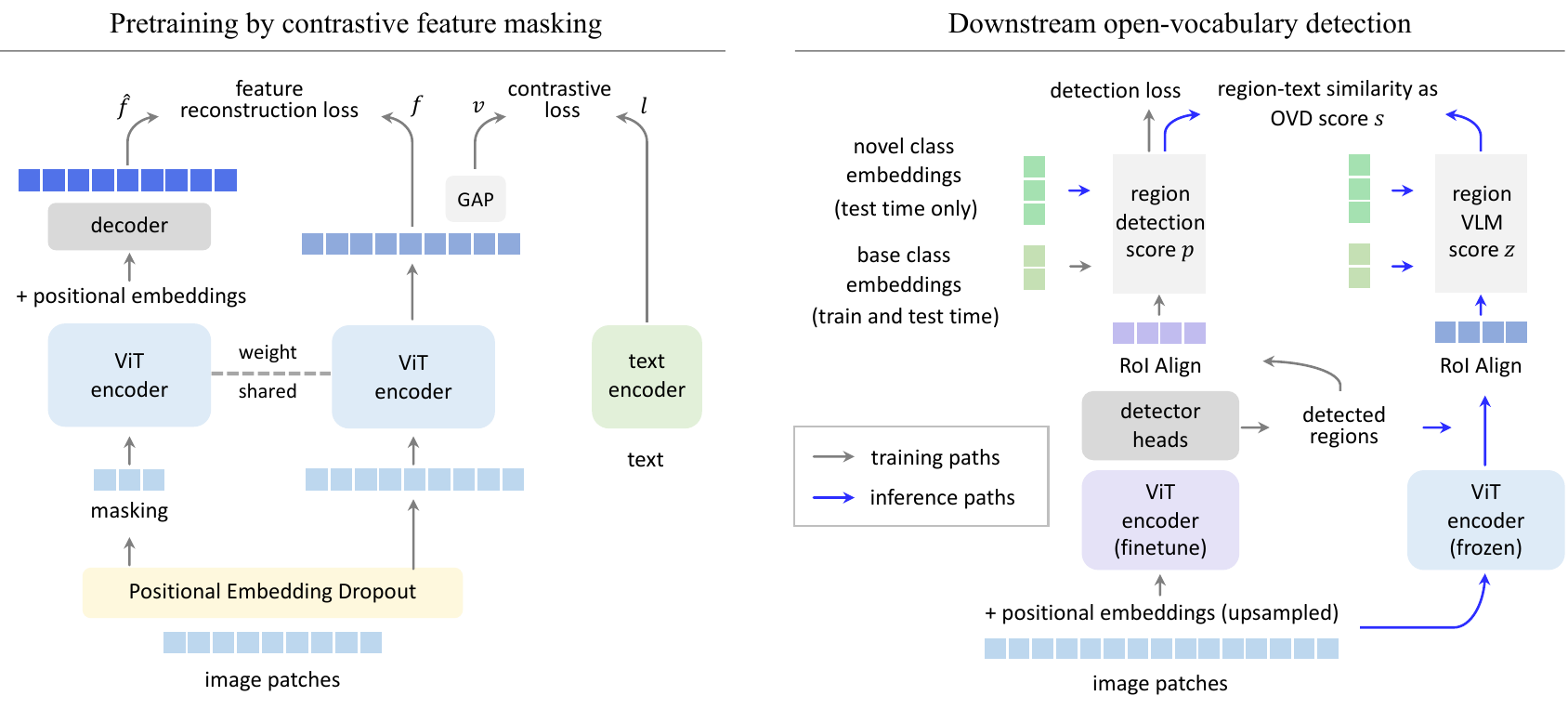}
\vspace{-1mm}
\caption{\textbf{\ours architecture:} We present both the image-text pretraining (left) and open-vocabulary detection finetuning (right) architecture of \ours. (Left) Building upon contrastive learning, we learn to reconstruct the masked tokens in the joint image-text embedding space. In addition, we propose Positional Embedding Dropout (PED) which randomly masks out the whole PE during pretraining to mitigate overfitting to the low-res positional embeddings, thus adapting better to the high-res downstream detection task. (Right) The open-vocabulary detector is initialized with the pretrained ViT backbone during finetuning. The detected region embeddings match with the cached category embeddings to compute the region scores. At inference, we exploit the frozen ViT backbone to obtain the VLM score $z$, which is combined with the detection score $p$ into the open-vocabulary detection score $s$ (Best viewed in color). 
}
\vspace{-3mm}
\label{fig:overview}
\end{figure*}

\section{Method}
\label{sec:method}

We tackle the problem of open-vocabulary object detection. During training, the model can access the detection labels of base categories, but at the inference phase, it must be able to detect objects from a set of novel categories. To achieve this, we utilize pretrained vision and language models (VLMs) following previous works~\cite{gu2022openvocabulary,zhong2021regionclip,kuo2022f}. However, instead of taking off-the-shelf pretrained VLM, we demonstrate how to better pretrain VLMs with vision transformers~\cite{dosovitskiy2020image} for open-vocabulary detection.

\subsection{Preliminaries: Overall Pipeline} 
\label{sec:preliminaries}
\paragraph{Pretraining.}\quad
We adopt a dual-encoder image-text contrastive model widely used in existing works~\cite{radford2021clip,align}. The image embeddings $\{{v}\}$ and text embeddings $\{{l}\}$ are obtained by global average pooling at the last layers of image and text encoders. The cosine similarity of the embeddings in batch $B$, scaled by a learnable temperature $\tau$ are the input to the InfoNCE loss~\cite{oord2018representation,radford2021clip}. The image-to-text (I2T) contrastive loss is formulated as:
\begin{equation}\label{eqn:contrastive}
L_{\text{I2T}} = -{1 \over {B}} \sum_{i=1}^{B} \log({\text{exp}(v_{i}l_{i} / \tau) \over { \sum_{j=1}^{B} \text{exp}(v_{i} l_{j} / \tau)  }}).
\end{equation}
The text-to-image (T2I) contrastive loss is symmetrical with the I2T loss by exchanging the inner/outer summation loops. The total contrastive loss $L_{con}$ is obtained by $L_{con} = (L_{\text{I2T}} + L_{\text{T2I}}) / 2$.

\paragraph{Downstream open-vocabulary detection.}\quad
Our open-vocabulary detection algorithm follows existing works~\cite{Zareian_2021_CVPR,gu2022openvocabulary, kuo2022f, kim2023region}. At training, for each detected region $i$, its region embedding is the RoI-Align feature. The detection score $p_i$ is the cosine similarity between the region embedding and text embeddings of $C_B$ followed by a softmax. Note the text embeddings are computed from the same text encoder from the image-text pretraining. At test time, the text embeddings are expanded from the $C_B$ to $C_B \cup C_N$ plus the ``background" embedding. We also extract VLM embedding of region $i$ by RoI-Align at the last feature map of the ViT backbone. The VLM score $z_i$ is the cosine similarity with the $C_B \cup C_N$ text embeddings. Similarly, the detection score $p_i$ is now computed with $C_B \cup C_N$ text embeddings.

An object detector for open-vocabulary scenarios is trained on the labels of base categories $C_B$, but must be capable of detecting the union of base and novel categories ($C_B \cup C_N$) at test time. Following existing works~\cite{Zareian_2021_CVPR,gu2022openvocabulary}, we replace the fixed-size classifier layer with the text embeddings of base categories. The same text encoder from the image-text pretraining is used to compute the text embeddings to maintain the pretrained open-vocabulary knowledge. The ``background" phrase represents the background category, and the proposals not matched to any $C_B$ annotations are labeled as background. 

The ensemble open-vocabulary detection score ${s_i}^{\text{ens}}$ is obtained by geometric means~\cite{gu2022openvocabulary,kuo2022f}:
\begin{equation}\label{eqn:combine-score}
{s_i}^{\text{ens}} = \begin{cases}
    z_i^{(1-\alpha)} \cdot p_i^\alpha & \text{if } i \in C_B\\
    z_i^{(1-\beta)} \cdot p_i^\beta & \text{if } i \in C_N
\end{cases}
\end{equation}
, where $\alpha, \beta \in [0, 1]$ control the weights for base and novel categories. The background score comes directly from the detection score $p_i$, because the VLM score with ``background" phrase tends to be not as reliable.

\subsection{Contrastive Feature Masking}
\label{sec:method:pretraining}
Our method performs reconstruction in the joint image-text embedding space (see \figref{fig:overview}-left) as an auxiliary objective to the contrastive image-text learning (in \secref{sec:preliminaries}).

\paragraph{Masked feature reconstruction.}\quad
Following MAE~\cite{he2022masked}, we randomly mask a large portion of image tokens (\eg, mask ratio 75\%) for representation learning. However unlike MAE, we predict the joint image-text embedding instead of the raw pixels to encourage better learning of semantics. Specifically, the output features $\{f\}$ of the contrastive image encoder before the global average pooling is our reconstruction target. We use the cosine distance between the reconstructed features $\{\hat{f}\}$ and unmasked image features $\{f\}$ as loss function. Let $M$ be the set of masked patch indices, and our reconstruction loss $L_{rec}$ is computed only on the masked tokens as:
\begin{equation}\label{eqn:reconstruction}
L_{rec} = 1 - {1 \over {B}} \sum_{i=1}^{B}({1 \over {|M|}} \sum_{k\in M} {{f \cdot \texttt{sg}(\hat{f})} \over {{\|f\| \cdot \|\texttt{sg}(\hat{f})}\|}}),
\end{equation}
where $|M|$ is the number of masked tokens and \texttt{sg} denotes stop gradient. The total \ours loss is $L_{con} + L_{rec}$.

Our reconstruction encoder is identical (weight-shared) to the contrastive image encoder, but applied only on the visible, unmasked tokens (\eg, 25\%). The decoder takes the encoded visible tokens and learnable \texttt{[mask]} tokens added with positional embeddings.

\paragraph{Faster training by contrastive branch masking.}\quad
The feature reconstruction branch adds a computation burden (\eg 25\%) to the pretraining depending on the masking ratio (\eg 75\%). We note that this cost can be waived by feeding only the masked tokens ($M$) to the contrastive branch, so that the input patches to the contrastive and reconstruction encoders are mutually exclusive, and yields the same reconstruction target $\{\hat{f}_{k\in M}\}$. Our ablation study in Table~\ref{tab:ablation:tokenratio} shows that this technique maintains the training efficiency of contrastive learning, while still achieves significant gains over the baseline in open-vocabulary detection.

\paragraph{Positional embedding dropout.}\quad In vision transformer encoder, positional embeddings are added to all tokens after the first patchifying layer to provide the location of each patch in the image. While the positional embeddings work well for image classification/retrieval, it tends to overfit to the lower-resolution object-centric images, and struggle with higher-resolution images typically used by detection task. In addition, the recognition of objects in detection occurs at region- rather than image-level  (\eg see VLM scores $z_i$ for region $i$ in \secref{sec:preliminaries}), which causes difficulty for the positional embeddings trained only for image-level task.

We propose a simple yet effective technique called Positional Embedding Dropout (PED) to address this problem by randomly masking out the whole positional embeddings during training (\eg, with a probability 0.5). This teaches the model not to rely heavily on the positional embeddings and thus can process the high-res images and perform better region classification. PED not only outperforms both the baseline and `no positional embeddings' variants, but enables the use of \textit{frozen} vision transformer to achieve further improvement in open-vocabulary detection.

\subsection{Open-vocabulary Detection}
\label{sec:method:ovd}
An object detector for open-vocabulary scenarios is trained on the labels of base categories $C_B$, but must be capable of detecting the union of base and novel categories ($C_B \cup C_N$) at test time (see \secref{sec:preliminaries} and \figref{fig:overview}-right).

\paragraph{Baseline architecture.}\quad
Our detector adopts the simple feature pyramid and windowed attention to handle higher resolution images as proposed in ViTDet~\cite{li2022exploring}, and employs Mask R-CNN heads and class-agnostic box regression and mask heads as in~\cite{du2022learning,gu2022openvocabulary,Zareian_2021_CVPR,zhong2021regionclip,kuo2022f}. In addition, we leverage a recent novel object proposal method~\cite{kim2022learning} by replacing the binary classification in the RPN with the centerness-based objectness. The predicted objectness score $o_i$ is combined into the final OVD score as ${s_i}^{\text{OVD}} = {o_i} \cdot {s_i}^{\text{ens}}$.

Our detector backbone is initialized with the pretrained ViT in the VLM from \secref{sec:method:pretraining}, and is finetuned together with the newly added detector heads. Note we do \textit{not} apply positional embedding dropout (PED) during finetuning as the location information is critical in detection.

\paragraph{Backbone learning rate.}\quad
As the pretrained knowledge in the backbone is critical in recognizing novel categories, it is important to set the backbone learning rate so as to prevent forgetting in the finetuning phase. On the other hand, entirely freezing the backbone limits the ability to adapt to detection tasks. We find that setting the backbone learning rate lower (\eg, 0.5$\times$) than the rest of the detector layers shows advantage in the trade-off. After the detection training is done, we explore using the frozen ViT backbone at test time, as described next.

\begin{table}[t]
\centering
\small
\tablestyle{6.0pt}{1.12}
\begin{tabular}{l|ll|ll}
\multirow{2}{*}{method} & pretrained& detector & \multirow{2}{*}{\bf{AP$_r$}} & \multirow{2}{*}{\gray{AP}} \\
                      & model     & backbone  & & \\
\shline
\textbf{ConvNet based:} & & & & \\
DetPro-Cascade~\cite{du2022learning}      & ViT-B/32  & R-50          & 20.0      & \gray{27.0} \\
Detic-CN2~\cite{zhou2022detecting}           & ViT-B/32  & R-50          & 24.6      & \gray{32.4} \\
RegionCLIP~\cite{zhong2021regionclip}          & R-50x4    & R-50x4        & 22.0      & \gray{32.3} \\
ViLD-Ens~\cite{gu2022openvocabulary}            & ViT-B/32  & R-152         & 18.7      & \gray{26.0} \\
ViLD-Ens~\cite{gu2022openvocabulary}            & ViT-L/14  & EffNet-B7     & 21.7      & \gray{29.6} \\
ViLD-Ens~\cite{gu2022openvocabulary}            & EffNet-B7 & EffNet-B7     & 26.3      & \gray{29.3} \\
VL-PLM~\cite{zhao2022exploiting}              & ViT-B/32  & R-50          & 17.2      & \gray{27.0} \\
OV-DETR~\cite{zang2022open}            & ViT-B/32 & R-50     & 17.4      & \gray{26.6} \\
Rasheed~\etal~\cite{rasheed2022bridging}     & ViT-B/32  & R-50          & 21.1      & \gray{25.9} \\
PromptDet~\cite{feng2022promptdet}            & ViT-B/32 & R-50     & 21.4      & \gray{25.3} \\

\hline
\textbf{ViT based:} & & & & \\
OWL-ViT~\cite{minderer2022simple}             & ViT-H/14  & ViT-H/14      & 23.3$^*$     & \gray{35.3$^*$}  \\
OWL-ViT~\cite{minderer2022simple}              & ViT-L/14  & ViT-L/14      & 25.6$^*$     & \gray{34.7$^*$} \\
{\bf{\ours (ours)}}      & ViT-B/16  & ViT-B/16      & \bf{29.6}$^*$& \gray{33.8$^*$} \\
{\bf{\ours (ours)}}       & ViT-L/16  & ViT-L/16      & \bf{35.6}$^*$& \gray{38.5$^*$} \\
\arrayrulecolor{lightgray}\hline\arrayrulecolor{black}
\bf{\ours (ours)}    & ViT-B/16  & ViT-B/16      & \bf{28.8} & \gray{32.0} \\
\bf{\ours (ours)}    & ViT-L/16  & ViT-L/16      & \bf{33.9} & \gray{36.6} \\
\end{tabular}
\vspace{-0mm}
\caption{\textbf{LVIS open-vocabulary object detection.} \ours outperforms the best existing approach by {+7.6 AP$_r$}, and the other ViT-based approach~\cite{minderer2022simple} by {+10.0 AP$_r$} using the same backbone. 
$*$: reports box AP.
}
\vspace{-4mm}
\label{tab:ovd_lvis}
\end{table}

\paragraph{Frozen backbone inference}\quad
\label{sec:method:frozen}
While the ViT backbone adapts to the detection tasks, it tends to forget some of the pretrained open-vocabulary knowledge. Therefore, for inference, we propose to use a separate frozen ViT backbone as an open-vocabulary region classifier.
Specifically, we use the frozen backbone instead of the finetuned backbone when computing the region VLM score $z_i$ (\secref{sec:preliminaries}). We find it important for the frozen ViT to be pretrained with our positional embedding dropout (PED), to serve as a strong zero-shot region classifier. We show by experiments that incorporating the PED pretraining and frozen backbone inference provides large gains in open-vocabulary detection.

\begin{table}[t]
\centering
\small
\tablestyle{5.0pt}{1.12}
\begin{tabular}{l|ll|cc}
\multirow{2}{*}{method} & pretrained & detector & \multirow{2}{*}{\bf{novel AP}} & \multirow{2}{*}{\gray{AP}} \\
                      & model     & backbone  & & \\
\shline
\textbf{ConvNet based:} & & & & \\
ViLD~\cite{gu2022openvocabulary}            & ViT-B/32  & R-50   & 27.6  & \gray{51.3} \\
OV-DETR~\cite{zang2022open}            & ViT-B/32  & R-50   & 29.4  & \gray{52.7} \\
\textbf{w/ pseudo box labels:} & & & & \\
XPM~\etal~\cite{huynh2022open}        & R-50  & R-50   & 27.0   & \gray{41.2} \\
RegionCLIP~\cite{zhong2021regionclip} $\dagger$  & R-50x4    & R-50x4     & 39.3  & \gray{55.7} \\
PromptDet~\cite{feng2022promptdet}          & ViT-B/32  & R-50   & 26.6  & \gray{50.6} \\
VL-PLM~\cite{zhao2022exploiting}            & ViT-B/32  & R-50   & 34.4  & \gray{53.5} \\
Rasheed~\etal~\cite{rasheed2022bridging} $\ddagger$   & ViT-B/32  & R-50   & 36.9  & \gray{51.5} \\
\textbf{w/ weak supervision:} & & & & \\
Detic-CN2~\cite{zhou2022detecting}          & ViT-B/32  & R-50          & 24.6      & \gray{32.4} \\
\hline
\textbf{ViT based:*} & & & & \\
\bf{\ours (ours)}    & ViT-B/16  & ViT-B/16      & {30.8} & \gray{42.4} \\
\bf{\ours (ours)}    & ViT-L/16  & ViT-L/16      &  {34.1} & \gray{46.0} \\
\end{tabular}
\vspace{-1mm}
\caption{{\textbf{COCO open-vocabulary object detection (box AP50).}} \ours represents the first ViT-based approach and demonstrates a very competitive novel AP without using pseudo labeling or weak supervision. $\dagger$: RegionCLIP uses an off-the-shelf RPN during its pretraining. $\ddagger$: Rasheed~\etal uses an external MViT detector~\cite{maaz2022class} during pretraining. *: The other ViT-based method~\cite{minderer2022simple} report their results on LVIS only.
}
\vspace{-6mm}
\label{tab:ovd_coco}
\end{table}

\section{Experimental Results}
\label{sec:results}

\paragraph{Pretraining setup.}\quad
For the image-text pretraining, we use the widely-used ViT-B/16 and ViT-L/16 as the image encoder, with an input image size of 224. We use the fixed 2D sinusoidal positional embeddings, and apply Positional Embedding Dropout (PED) with a drop probability of 0.5. The image embedding is obtained by global average pooling at the last ViT layer. The text encoder is a 12-layer Transformer as in~\cite{radford2021clip, yu2022coca}, with the input sequences truncated to a fixed length of 64 tokens. The L2-normalized image and text embeddings and a learnable scaling temperature are the input to the InfoNCE contrastive loss~\cite{radford2021clip}.

Our feature reconstruction decoder is a 2-layer ViT, unlike the 8-layer counterpart of MAE~\cite{he2022masked} designed for raw pixel reconstruction. The reconstruction loss is cosine distance, scaled by a loss coefficient 2.0, and is added to the contrastive loss. We use ALIGN dataset~\cite{align} by default, while we show using LAION datasets~\cite{schuhmann2021laion} leads to similar results (\tabref{tab:pretraining_data}). Unless noted, we use a batch size of 4k for ablation and 16k for comparisons, and train for 500k iterations using the AdamW optimizer with an initial learning rate (LR) of 5e-4 and linear LR decay. We use 10k warm-up iterations and a weight decay of 0.01.

\paragraph{Detection finetuning setup.}\quad
We train our model on base categories $C_B$ with an image size of 1024$\times$1024. The positional embeddings (PE) are bilinearly interpolated to fit the higher resolution. We do \textit{not} apply PE Dropout during the detection training, and set a lower learning rate for the backbone (\eg, 0.5 $\times$) compared to the rest of the model. We utilize CLIP templates~\cite{radford2021clip} and take the average text embeddings of each category. We use a batch size 128, the SGD optimizer with momentum 0.9, an initial learning rate of 0.18/0.02 and train for 36.8k/11.3k iterations on LVIS/COCO datasets.

\begin{table*}[t]
\centering
\small
\tablestyle{6pt}{1.1}
\begin{tabular}{l|c|cccccc|cccccc}
&  image & 
\multicolumn{6}{c|}{Flickr30K (1K test set)} & \multicolumn{6}{c}{MS COCO (5K test set)} \\
& encoder & \multicolumn{3}{c}{\underline{{\white{-------}}image-to-text{\white{-------}}}} & \multicolumn{3}{c|}{\underline{{\white{-------}}text-to-image{\white{-------}}}} & \multicolumn{3}{c}{\underline{{\white{-------}}image-to-text{\white{-------}}}} & \multicolumn{3}{c}{\underline{{\white{-------}}text-to-image{\white{-------}}}} \\
method & size      & R@1  & R@5  & R@10 & R@1  & R@5   & R@10      & R@1  & R@5  & R10 & R@1  & R@5 & R@10    \\
\shline
CLIP~\cite{radford2021clip}   & 302M
      & 88.0 & 98.7 & 99.4 & 68.7 & 90.6 & 95.2        & 58.4 & 81.5 & 88.1 & 37.8 & 62.4 & 72.2 \\
ALIGN~\cite{align}   & 480M
      & 88.6 & 98.7 & 99.7 & 75.7 & 93.8 & 96.8        & 58.6 & 83.0 & 89.7 & 45.6 & 69.8 & 78.6 \\
FLAVA~\cite{singh2022flava}   & 86M
        & 67.7 & 94.0 & - & 65.2 & 89.4 & -             & 42.7 & 76.8 & - & 38.4 & 67.5 & -     \\
FILIP~\cite{yao2021filip}   & 302M
      & 89.8 & 99.2 & {99.8} & 75.0 & 93.4 & 96.3       & 61.3 & 84.3 & 90.4 & 45.9 & 70.6 & 79.3 \\
Florence~\cite{yuan2021florence} & 637M
         & 90.9 & 99.1 & - & 76.7 & 93.6 & -             & 64.7 & 85.9 & - & 47.2&  71.4&  -    \\
CoCa-Large~\cite{yu2022coca} & 303M
 & 91.4 & \bf{99.2} & \bf{99.9} & 79.0& 95.1 & 97.4      & 65.4 & 85.6 & {91.4}&  \bf{50.1} & \bf{73.8} & \bf{81.8}  \\
\hline
\bf{\ours (ours)}   & 303M & \bf{91.7}  & {99.0}  & \bf{99.9}   & \bf{79.6}  & \bf{95.6}  & \bf{97.7} 
& \bf{66.4}  & \bf{86.1}  & \bf{91.5}  & {49.8}  & {73.5}  & {81.6}
         \\ 
\end{tabular}
\caption{\textbf{Zero-shot image-text retrieval results on Flickr30K and COCO benchmarks.} We evaluate our pretrained model compared to other methods.
We outperform the state-of-the-art CoCa-Large with the same backbone in 8 out of 12 metrics.}
\label{tab:retrieval_sota}
\vspace{-2mm}
\end{table*}

\subsection{Main Results}
\label{sec:exp:ovd}
\paragraph{LVIS benchmark.}\quad
We compare with other methods on the LVIS~\cite{lvis} open-vocabulary detection benchmark which contains a diverse set of 1203 object categories. The base categories $C_B$ for training are the `frequent' and `common' categories, and novel categories $C_N$ are the `rare' categories which are held out for testing, as in~\cite{gu2022openvocabulary,zhong2021learning,du2022learning}. The main metric is mask AP$_r$, and we report the mean over three runs following~\cite{gu2022openvocabulary} for reproducibility.

\tabref{tab:ovd_lvis} reports that the best \ours model achieves 33.9 AP$_r$, a significant improvement over the best existing ViT-based method OWL-ViT~\cite{minderer2022simple} by {+10.0} AP$_r$. Remarkably, \ours using a smaller ViT-B/16 backbone outperforms OWL-ViT with ViT-L/14 by {+4.0} AP$_r$. Furthermore, compared to the current best approach ViLD-Ens with EffNet-B7 backbone, \ours achieves a {+7.6} AP$_r$ improvement. Notably, \ours has a simple finetuning recipe using only vanilla detection losses~\cite{he2017mask}, without the use of long-tail recognition losses~\cite{minderer2022simple,zhong2021regionclip,zhou2022detecting}, knowledge distillation~\cite{gu2022openvocabulary,du2022learning}, weak supervision~\cite{zhou2022detecting}, or pseudo box/mask labels~\cite{zhong2021regionclip,zhao2022exploiting,rasheed2022bridging}, all of which are common among current open-vocabulary detection methods.

\paragraph{COCO benchmark.}\quad
We present the comparison on the COCO open-vocabulary detection benchmark. This setup uses 48 base categories for training and 17 novel categories for testing~\cite{gu2022openvocabulary}. The main metric is AP50 of novel categories (`novel AP'). Due to fewer training categories, the \ours model has a tendency to overfit to these categories using only the vanilla detection losses. This is because \ours do not use any auxiliary objectives such as pseudo box/mask labels~\cite{huynh2022open,feng2022promptdet,zhong2021regionclip,zhao2022exploiting,rasheed2022bridging}, knowledge distillation~\cite{gu2022openvocabulary,du2022learning}, weak supervision~\cite{zhou2022detecting} to counter-balance overfitting on this benchmark. However, \tabref{tab:ovd_coco} shows that \ours is still very competitive among existing methods leveraging auxiliary objectives. Moreover, \ours represents the first ViT-based method on this benchmark, as the other ViT-based~\cite{minderer2022simple} approach only benchmarks on LVIS.

\begin{table}[t]
\centering
\small
\tablestyle{8pt}{1.1}
\begin{tabular}{ll|ccc}
method              & backbone  & AP    & AP\textsubscript{50}  & AP\textsubscript{75}    \\
\shline
\gray{supervised~\cite{gu2022openvocabulary}} & \gray{R-50} & \gray{25.6} & \gray{38.6} & \gray{28.0}  \\
ViLD~\cite{gu2022openvocabulary}   & R-50      & 11.8      & 18.2      & 12.6  \\
DetPro~\cite{du2022learning}       & R-50      & 12.1      & 18.8      & 12.9  \\
\hline
\bf{\ours (ours)}   & ViT-B/16  & \bf{15.9}    & \bf{24.6}    & \bf{17.4} \\
\bf{\ours (ours)}   & ViT-L/16  & \bf{18.7}    & \bf{28.9}    & \bf{20.3} \\
\end{tabular}
\caption{\textbf{Transfer detection on Objects365 (Box APs).} All models are trained on the LVIS base categories and tested on Objects365 dataset, without finetuning.}
\label{tab:transfer}
\vspace{-2mm}
\end{table}

\paragraph{Zero-shot Image-Text Retrieval.} \label{sec:retrieval}
In addition to our main evaluation on the region-level open-vocabulary detection, we evaluate our \textit{image-level} representation in zero-shot image-text retrieval. We take the same \ours model as in the last row of \tabref{tab:ovd_lvis} (ViT-L, batch size 16k) and continue the pretraining on higher resolution, \eg, 448, for extra 40K iterations, following the standard protocol~\cite{align,yu2022coca}.

\begin{table*}[t]
\centering
\subfloat[\footnotesize{\textbf{Masked reconstruction}. `baseline' is the contrastive image-text pretraining. Our proposed masked feature reconstruction improves by +3.3 AP$_r$. Reconstruction in the raw pixel space or the first-layer feature space shows no benefit.\label{tab:ablation:pretraining}}]{
\tablestyle{5.0pt}{1.1}\begin{tabular}{lcc}
{pretraining method}  & AP$_r$  &   \gray{AP} \\
\shline
baseline                 & 27.4 {\white{(+x.x)}}    & \gray{30.4}  \\
w/ feat recon.           & 30.7 ({\blue{+3.3}})     & \gray{34.0}  \\
\hline
w/ pixel recon.          & 27.1  {\white{(+x.x)}}   & \gray{31.3} \\
w/ 1st-layer feat recon. & 27.2  {\white{(+x.x)}}   & \gray{30.8}  \\
 \multicolumn{3}{c}{~}\\
\end{tabular}}\hspace{6mm}
\subfloat[\footnotesize{\textbf{Positional Embedding Dropout (PED)} improves the baseline by 1.1 AP$_r$. It achieves a further gain of +2.7 when used with masked feature reconstruction. PED outperforms `no PE' by 3.5 / 1.6 AP$_r$ with/without feature reconstruction\label{tab:ablation:ped}}]{
\tablestyle{5.0pt}{1.1}\begin{tabular}{lcc}
{pretraining method}  & AP$_r$  &   \gray{AP} \\
\shline
baseline                 & 27.4 {\white{(+x.x)}}   & \gray{30.4}  \\
w/ PED                   & 28.5 ({\blue{+1.1}})    & \gray{31.9} \\
w/ feat recon. + PED     & 31.2 ({\blue{+3.8}})    & \gray{33.7} \\
\hline
w/ no PE                & 25.8 {\white{(+x.x)}}     & \gray{29.5}  \\
w/ feat recon. + no PE  & 27.7 {\white{(+x.x)}}     & \gray{31.9}  \\
\end{tabular}}\hspace{6mm}
\subfloat[\footnotesize{\textbf{Masking contrastive branch} recovers the training efficiency with little or no performance drop, outperforming the baseline by +3.0 AP$_r$.\label{tab:ablation:tokenratio}}]{
\tablestyle{5.5pt}{1.1}\begin{tabular}{llll}
contr. / recon.            & FLOPs &   AP$_r$  &   \gray{AP}\\
\shline
100\% / {\white{0}}0\%    & 1.00$\times$ & 27.4    & \gray{30.4}   \\
100\% / 25\%              & 1.23$\times$ & 30.7    & \gray{34.0}   \\
100\% / 50\%              & 1.44$\times$ & 29.9   & \gray{33.1}   \\
\hline
{{\white{0}}75\% / 25\%}
                          & \textbf{1.01}$\times$  & 30.4   & \gray{33.9}    \\
 \multicolumn{4}{c}{~}\\
\end{tabular}}
\vspace{4mm}
\subfloat[\footnotesize{\textbf{Backbone fine- tuning lr ratio} (bblr) w.r.t. added detector layers.\label{tab:ablation:bblr}}]{
\tablestyle{4.0pt}{1.1}\begin{tabular}{lcc}
bblr &   AP$_r$  &   \gray{AP} \\
\shline
0.0          & 9.5       & \gray{11.4}  \\
0.1          & 25.8      & \gray{28.5}  \\
0.5          & \bf{27.4} & \gray{30.4}  \\
1.0          & 26.0      & \gray{30.2}  \\
\end{tabular}}\hspace{4mm}
\subfloat[\footnotesize{\textbf{Frozen backbone inference.} When using standard positional embeddings, it underperforms the finetuned encoder. In contrast, with the encoder pretrained with PED, the frozen backbone inference surpasses the finetuned counterpart by +2.0 and +1.3 AP$_r$.\label{tab:ablation:frozen}}]{
\tablestyle{4pt}{1.1}\begin{tabular}{lccc}
 & w/ PED & AP$_r$  &   \gray{AP} \\
\shline
baseline                    & & 27.4 {\white{-}}$\rightarrow${\white{-}} 24.6 {\white{-}}({\red{-2.8}}) & \gray{30.4 $\rightarrow$ 30.3 }  \\
w/ feat-recon.              & & 30.7 {\white{-}}$\rightarrow${\white{-}} 27.1 {\white{-}}({\red{-3.8}}) & \gray{34.0 $\rightarrow$ 33.4}  \\
\hline
baseline    & \checkmark    & 28.5 {\white{-}}$\rightarrow${\white{-}} 30.5 {\white{-}}({\blue{+2.0}}) & \gray{31.9 $\rightarrow$ 31.8}  \\
w/ feat-recon  & \checkmark    & 31.2 {\white{-}}$\rightarrow${\white{-}} 32.5 {\white{-}}({\blue{+1.3}}) & \gray{33.7 $\rightarrow$ 34.1}  \\
\end{tabular}}\hspace{4mm}
\subfloat[\footnotesize{\textbf{Scalabiltiy:} The benefit of \textbf{`baseline $\rightarrow$ \ours'} across different model and contrastive batch sizes. It improves the baselines by +2.4 to +5.1 AP$_r$.\label{tab:ablation:scaling}}]{
\tablestyle{4.0pt}{1.1}\begin{tabular}{cccc}
model & batch & AP$_r$  &   \gray{AP} \\
\shline
B/16 & 4k      & 24.1 {\white{-}}$\rightarrow${\white{-}} 26.8 {\white{-}}({\blue{+2.7}}) & \gray{27.6 $\rightarrow$ 30.2 }  \\
B/16 & 16k      & 26.4 {\white{-}}$\rightarrow${\white{-}} 28.8 {\white{-}}({\blue{+2.4}}) & \gray{30.3 $\rightarrow$ 33,5}  \\
L/16 & 4k      & 27.4 {\white{-}}$\rightarrow${\white{-}} 32.5 {\white{-}}({\blue{+5.1}}) & \gray{30.4 $\rightarrow$ 34.1}  \\
L/16 & 16k      & 30.5 {\white{-}}$\rightarrow${\white{-}} 33.9 {\white{-}}({\blue{+3.4}}) & \gray{35.9 $\rightarrow$ 36.6}  \\
\end{tabular}}
\caption{\textbf{Ablation studies} on LVIS open-vocabulary detection benchmark. We train on base (`frequent' + `common') categories, test on novel (`rare') categories, and report AP$_r$. We use ViT-L/16 backbone and contrastive batch size 4k unless otherwise noted.}
\label{tab:ablations}
\vspace{-2mm}
\end{table*}

\tabref{tab:retrieval_sota} shows our comparison with other dual-encoder methods on Flickr30K and MS COCO benchmarks. \ours outperforms state-of-the-art methods of similar or larger model size, on 8 out of 12 metrics.

\paragraph{Zero-shot Transfer Detection.} \label{sec:exp:transfer}
To assess \ours's ability to generalize in zero-shot transfer detection, we test its performance on Objects365-v1 validation split~\cite{objects365}. We use the same detector trained on LVIS base categories (\tabref{tab:ovd_lvis}) and replace LVIS with Objects365 vocabulary embeddings for transfer detection without finetuning~\cite{gu2022openvocabulary,du2022learning}. We assume all categories are novel and set $\alpha, \beta$=(0.0, 0.65) in \eqnref{eqn:combine-score}. Our best model achieves 18.7 AP, outperforming ViLD by +6.9 AP and DetPro by +5.6 AP, as shown in \tabref{tab:transfer}. Given the different backbone capacity (R50 vs ViT), this comparison mainly serves to demonstrate that \ours can achieve strong cross-dataset generalization.

\vspace{-1mm}
\subsection{Ablation Study}
We ablate the design of \ours's pretraining and open-vocabulary detector. We evaluate on the LVIS open-vocabulary detection benchmark. The image encoder is ViT-L/16, and contrastive batch size is 4k by default.

\paragraph{Masked feature reconstruction.}\quad
\tabref{tab:ablation:pretraining} ablates the proposed masked image-text pretraining (\secref{sec:method:pretraining}). The proposed masked feature reconstruction offers a clear benefit of +3.3 AP$_r$ over the contrastive image-text pretraining baseline. In this case, the feature reconstruction target is the output features of the image encoder. We compare with other reconstruction targets: normalized image pixels~\cite{he2022masked} and the features from the first patchifying layer. We observe that neither improve over the baseline, likely because the contrastive pretraining sets a strong baseline representation~\cite{gu2022openvocabulary,dong2022clip,kuo2022f}. In contrast, the proposed masked feature reconstruction clearly improves upon the strong baseline and shows advantage in open-vocabulary detection.

\paragraph{Positional embedding dropout.}\quad
In \tabref{tab:ablation:ped}, we ablate the positional embedding dropout (`PED'). PED brings a gain of +1.1 AP$_r$ over the baseline (PE without dropout). This shows that PED effectively reduces overfitting to the low-res whole-image PE during pretraining, thus adapting better to the high-res detection task through finetuning. In addition, PED achieves further gain of +2.7 when used together with masked feature reconstruction. We compare PED with another baseline which uses no positional embeddings in the ViT encoder (`no PE'). The PED method outperforms the `no PE' baseline by 3.5 / 1.6 AP$_r$ with/without feature reconstruction. We note that the positional embeddings in the reconstruction decoder~\cite{he2022masked} is always kept. Finally, PED allows the use of the \textit{frozen} backbone as a strong \textit{region} classifier as shown in \tabref{tab:ablation:frozen}.

\paragraph{Faster training by masking contrastive branch.}\quad
\tabref{tab:ablation:tokenratio} studies image masking ratios of the contrastive and reconstruction encoders. By default, we apply our contrastive encoder on intact images during training, \ie 100$\%$ tokens. Adding the reconstruction tower with 25$\%$ input tokens results in 1.23$\times$ more training cost. To maintain the training efficiency, we explore feeding only 75$\%$ tokens to the contrastive encoder that are mutually exclusive from the reconstruction branch inputs. This masking technique fully recovers the training efficiency with little or no accuracy loss, outperforming the baseline by +3.0 AP$_r$.

\paragraph{Backbone learning rate ratio.}\quad
\ours requires the retention of pretrained knowledge in the backbone to recognize novel categories. \tabref{tab:ablation:bblr} reports the advantage to set the backbone learning rate lower than the rest of the detector during the finetuning, with a ratio 0.5$\times$ being the optimal value. Higher ratios lead to forgetting, while lower ratios limit the ability to adapt to the detection task.

\begin{table}[t]
\centering
\tablestyle{10.0pt}{1.1}\begin{tabular}{lcc}
{pretraining data}  & AP$_r$  &   \gray{AP} \\
\shline
ALIGN~\cite{align}   & 32.5    & \gray{34.1}  \\
LAION-2B~\cite{schuhmann2021laion}     & 32.4    & \gray{34.3} \\
LAION-400MB~\cite{schuhmann2021laion}  & 32.2   & \gray{34.1} \\
\end{tabular}
\vspace{-1mm}
\caption{\textbf{Pretraining data.} ViT-L/16 and batch size 4k is used.}
\label{tab:pretraining_data}
\vspace{-2mm}
\end{table}

\begin{table}[t]
\centering
\tablestyle{7pt}{1.1}\begin{tabular}{lcccc}
 &
\multicolumn{2}{c}{Flickr30K {\white{-}}} & \multicolumn{2}{c}{{\white{-}} MS COCO}\\
 & I2T & T2I {\white{-}} & {\white{-}} I2T & T2I \\
\shline
baseline        & 86.0 & 72.3 {\white{-}}&{\white{-}} 59.3 & 43.4 \\
w/ PED         & 86.1 & 72.5 {\white{-}}&{\white{-}} 59.1 & 43.2  \\
w/ feat recon. + PED & 87.0 & 73.6 {\white{-}}&{\white{-}} 60.1 & 44.2 \\
\end{tabular}
\caption{\textbf{Pretraining evaluation on zero-shot image-text retrieval (Recall@1).} We evaluate the image-level representation of our pretrained model on Flickr30k and COCO retrieval tasks. We ablate the positional embedding dropout (PED) and adding masked feature reconstruction. ViT-L/16 and batch size 4k is used.}
\label{tab:retrieval}
\vspace{-3mm}
\end{table}

\begin{figure*}[t]
\centering
\includegraphics[width=1.0\linewidth]{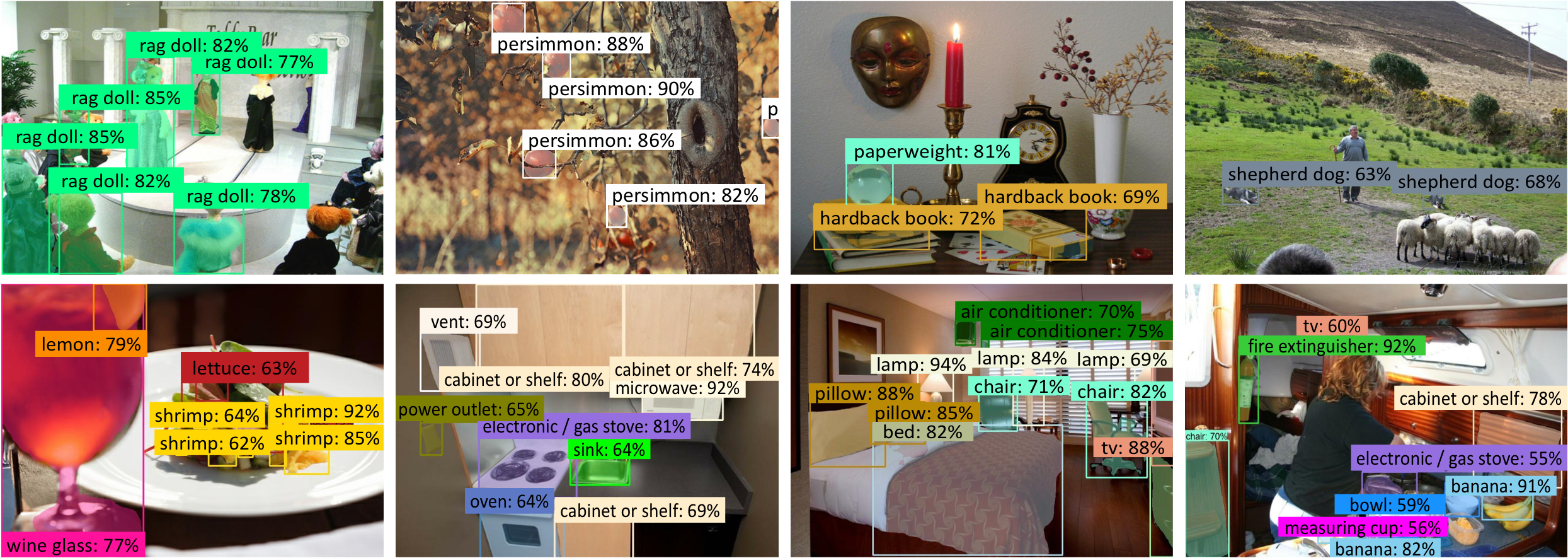}
\caption{\textbf{Qualitative results on LVIS novel categories (top) and Objects365 zero-shot transfer detection (bottom).} For LVIS results, we only show the novel categories for clarity. \ours detects many novel categories such as \textit{rag doll, persimmon, paperweight, hardback book, shepherd dog} on LVIS, and \textit{shrimp, power outlet} on Objects365.}
\label{fig:results}
\vspace{-3mm}
\end{figure*}

\paragraph{Frozen backbone inference.}\quad
Our ablation studies so far do \textit{not} involve frozen backbone inference. All ablations use the \textit{finetuned} ViT backbone to compute the VLM scores ($p_i$ in \secref{sec:preliminaries} and \eqnref{eqn:combine-score}). In \tabref{tab:ablation:frozen}, we assess the efficacy of the frozen backbone as a region classifier by substituting the finetuned ViT encoder with a frozen ViT encoder and analyze the performance (see the rightmost part of \figref{fig:overview}). Our experiments show that the frozen backbone underperforms the finetuned encoder when using standard positional embeddings, which applies to both the baseline with and without feature reconstruction loss. However, we find that pretraining the ViT encoder with positional embedding dropout (PED) leads to signficantly improved performance with frozen backbone, surpassing thoese of the finetuned backbone by +2.0/+1.3 AP$r$, without/with feature reconstruction loss. This result demonstrates the efficacy of PED in reducing the domain gap between contrastive pretraining and detection finetuning, thus improving zero-shot region classification. Combined with feature reconstruction, our full method achieves an overall improvement of +5.1 AP$r$ over the baseline.

\paragraph{Model size and batch size.}\quad
\tabref{tab:ablation:scaling} studies the effect of model size and batch size in \ours pretraining on the downstream open-vocabulary detection. We observe that increasing the batch size from 4k to 16k leads to an improvement of +2.7 / 1.4 AP$r$ for both ViT-B/L, while upgrading from ViT-B to ViT-L results in an improvement of +5.9 / 5.6 AP$r$ for both batch sizes. These results align with observations from the contrastive learning literature~\cite{radford2021clip,align,basic} that larger batch sizes and model sizes are both highly beneficial. Importantly, we find that \ours consistently outperforms the baseline by +2.4 to +5.1 AP$r$, across all batch and model sizes tested, further demonstrating its efficacy.

\paragraph{Pretraining data.}\quad
Apart from the ALIGN data~\cite{align}, we also experiment with LAION datasets~\cite{schuhmann2021laion} in \tabref{tab:pretraining_data}. LAION-2B / LAION-400M results in 32.4 / 32.2 AP$_r$, which is comparable to the ALIGN result 32.5 AP$_r$.

\paragraph{Image-text retrieval.}\quad
In addition to ablations on open-vocabulary detection, we investigate the effects of positional embedding dropout and masked feature reconstruction on zero-shot \textit{image-level} retrieval, and report the results in terms of Recall@1 metrics on Flickr30K and MS COCO datasets. \tabref{tab:retrieval} shows that positional embedding dropout effectively preserves the quality of image-level representation, while masked feature reconstruction yields an average improvement of 1\% Recall@1 across all metrics.

\vspace{-1mm}
\subsection{Visualizations}
\vspace{-1mm}
\paragraph{Feature reconstruction.}\quad
In \figref{fig:teaser}, we show our feature reconstruction results from our pretraining (\secref{sec:method:pretraining}). For visualization, we compute the similarity map (c) between the \textit{reconstructed} image features (d), and a query text embedding (e). We observe that the learned feature reconstructions are semantically plausible with respect to the queried image-text pairs.

\paragraph{Open-vocabulary detection outputs.}\quad
In \figref{fig:results}, we visualize our \ours outputs on LVIS novel categories (top row) and zero-shot transfer detection on Objects365 (bottom row). For both visualizations, we use the same model as in the last row of \tabref{tab:ovd_lvis}, which is trained on the LVIS base categories. On both datasets, \ours is able to detect many novel categories unavailable during training.

\vspace{-1mm}
\section{Conclusion}
\vspace{-1mm}

\label{sec:conclusion}
We introduce Contrastive Feature Masking Vision Transformer (\ours) which imbues the image-text pretraining with pixel/region-level semantics for open-vocabulary object detection. By using feature construction and positional embedding dropout, \ours is simple and scalable, outperforming the state-of-the-art on LVIS open-vocabulary detection benchmark by large margins, and shows very competitive performance on COCO benchmark and zero-shot transfer to Objects365. In addition, \ours outperforms the state-of-the-art on 8 out of 12 metrics of zero-shot image-text retrieval benchmarks. We hope \ours would inspire the community to explore image-text pretraining for open-vocabulary detection~\cite{kim2023region}.


{\small
\bibliographystyle{ieee_fullname}
\bibliography{egbib}
}

\end{document}